\documentclass[11pt,a4paper]{article}
\usepackage{float}
\usepackage[T1]{fontenc}
\usepackage{times}
\usepackage{lineno}
\usepackage{graphicx}
\usepackage{caption}
\usepackage{amsmath,amssymb}
\usepackage[margin=1in]{geometry}
\usepackage{authblk}
\usepackage{hyperref}
\usepackage{xurl}
\usepackage{natbib}
\usepackage{titlesec}
\usepackage{graphicx}
\usepackage{booktabs}
\bibpunct{}{ }{,}{s}{}{,}

\title{Large Language Models Threaten Double-blind Review}
\author[1]{Bulambo Mwendelwa Gloire}
\author[1,*]{Prasenjit Mitra}
\affil[1]{Carnegie Mellon University Africa, Kigali, Rwanda}
\date{}

\begin{document}
\maketitle

\section*{Abstract}
Double-blind peer review serves as the scientific community’s primary defense against status and affiliation bias. Its effectiveness rests on the assumption that anonymized manuscripts convey scientific merit without revealing their authors. While authorship can often be recovered using citation networks or stylistic markers, we show that this assumption is increasingly fragile in the presence of large language models (LLMs). Using only titles and abstracts from papers published after model training, we find that LLMs collapse anonymity more efficiently than humans, with belief concentrating onto a small subset of plausible authors drawn from pools of five domain-expert candidates. This vulnerability persists even when stylistic and bibliographic cues are excluded, indicating that stable patterns in problem framing and research focus function as latent conceptual signatures of authorship. Together, these findings indicate that double-blind review is vulnerable to automated semantic inference, necessitating a re-evaluation of how anonymity and fairness are maintained in an AI-augmented research ecosystem.

\section{Introduction}
Double-blind peer review is a cornerstone of efforts to promote fairness in scientific publishing \cite{ref4,ref20}. It is widely adopted in several disciplines, particularly in the social sciences, computer science, humanities, education, management and business research \cite{ref29,ref30}. By concealing author identities, it aims to mitigate biases linked to reputation, institutional affiliation, and prior visibility, allowing manuscripts to be judged on scientific merit alone \cite{ref17}. This system rests on a foundational assumption:
anonymized submissions do not contain sufficient information to reliably reveal their authors and thus they cannot be
easily unmasked. This assumption underlies anonymization guidelines and 
review policies across much of the scientific enterprise.

Prior research has shown that authorship can often be recovered when rich textual or structural information is available \cite{ref22, ref23}. Classical and modern attribution methods demonstrate that stylistic patterns, lexical regularities, and especially bibliographic and citation structures can encode strong identity signals \cite{ref4, ref6, ref19, ref20}. These findings have motivated increasingly strict anonymization practices focused on removing named entities, affiliations, and self-citations \cite{ref5, ref25}. However, most of such results rely on broad or weakly constrained candidate pools, where topical mismatches and residual metadata dominate inference \cite{ref10, ref18}. Does anonymization truly eliminate meaningful authorship signals when surface-level cues are removed and inference is limited to closely related domain experts?

Recent advances in large language models (LLMs) raise the possibility that anonymization is vulnerable at a deeper semantic level, and that deanonymization can be performed quickly and with minimal effort \cite{ref9, ref13, ref21}. While reviewers may not normally engage in elaborate efforts to identify authors, since actions that require significant time or effort are less likely to be undertaken, the widespread availability of LLMs at our fingertips lowers this barrier. With such tools readily accessible, a reviewer may perform a quick query to infer potential authorship within seconds. This easy availability may change the fundamental balance related to potential reviewer behavior.

Unlike earlier attribution approaches, modern LLMs encode scientific text in high-dimensional representations that capture fine-grained conceptual relationships. Identity leakage may therefore arise not from overt stylistic markers or bibliographic artifacts, but from latent regularities in how researchers frame problems, scope contributions, and position their work within a field \cite{ref27}. Existing studies of LLM-based authorship inference, however, are 
not reliable due to overt reliance on identifiable textual cues and the absence of clear temporal separation between model training data and evaluation samples \cite{ref16, ref28}.

In this work, we evaluate whether LLMs can deanonymize papers' authors given a few hunches that the reviewer may have about who the author may be. Using only titles and abstracts from papers published after model training data cutoff dates \cite{ref1, ref2, ref14, ref26}, and restricting inference to small sets of five plausible domain experts, we find that anonymity cannot be maintained. Where human researchers achieved only 11\% top-1 accuracy, large language models reached 42\% under identical conditions, representing a nearly fourfold improvement. This performance gap suggests that even among closely related domain expert candidates, minimal scientific text contains persistent authorship signals that human reviewers cannot reliably detect but which language models can exploit, posing a threat to the anonymity guarantees of double-blind peer review. These findings indicate that double-blind review faces structural limitations in the presence of modern language models, motivating a re-examination of how anonymity and fairness are preserved in contemporary peer review.

\section{Results} 
We analyzed a corpus of more than 12,000 scientific papers sourced from the Semantic Scholar API \cite{ref31}, spanning broad fields of study and published between 2024 and 2025.
We chose LLMs trained before this time period to ensure that
the LLM was not regurgitating the author names of a paper from rote memory ensuring no overlap with the documented training data of the evaluated models \cite{ref1,ref2}. 
In each trial, inference was performed on a candidate pool of five authors, four candidates constructed in two ways: one drawn from a large pool of random authors obtained via the Semantic Scholar API \cite{ref31}, and another from domain experts authors obtained via the Semantic Scholar Recommendation API \cite{ref3} and OpenAlex API \cite{ref32}, with exactly one of the candidates being the true author.
For each paper, we consider multiple definitions of the true author: (i) the last author, used as our primary baseline; (ii) the most senior author, and (iii) the most junior author. This design ensures that our findings are not tied to a single, potentially discipline-specific notion of authorship importance.
The model ranked likely authors among candidates and assigned log-probabilities interpreted as relative belief scores that is aggregated into confidence-aware suspect sets \emph{defined as the minimal set of candidates whose cumulative probability score exceeds a given threshold} $\tau$. The said threshold \emph{ is a confidence level representing the desired proportion of total probability mass to be covered}. For example, with a probability threshold $\tau$ of 0.9, the suspect set includes the smallest group of authors whose combined confidence score reaches 90\%, ensuring the true author is likely included. This formulation captures not only whether the true author is prioritized, but how rapidly effective anonymity collapses as confidence increases.

\subsection{Human authorship inference reveals limited selectivity}
We begin by evaluating human performance in inferring authorship from minimal scientific content, restricted to titles and abstracts. To establish a meaningful human baseline, we recruited twenty graduate researchers from a globally ranked institution embedded within an international network of research affiliations. All participants were active researchers with several years of experience conducting and publishing research in machine learning and related fields, affiliated with diverse institutions across both academic and industry settings. Their domain expertise ensured that the human evaluation reflected the judgment of informed, research active scientists, providing a strong and informed human baseline against which model performance could be assessed.

They participated in a closed-set probabilistic attribution task  in which each participant evaluated five papers from the machine learning literature, presented as title-abstract pairs. For each paper, participants were shown a candidate set of five authors comprising one true author and four domain-expert distractors, selected using the same procedure applied in the model evaluation. Participants assigned confidence scores to each candidate, reflecting their belief that the candidate authored the paper. These scores were normalized post hoc to form probability distributions over the candidate set, enabling computation of top-k accuracy, mean reciprocal rank (MRR), and confidence-aware suspect sets under varying thresholds $\tau$.

Human performance is limited in precise attribution. The true author is ranked first in only 11\% of cases (Table~\ref{tab:topk_accuracy_table}). The ranking quality remains low, with an overall mean reciprocal rank (MRR) of 0.36 (95\% CI: [0.31, 0.41]), indicating that correct authors are not consistently prioritized. At a moderate confidence threshold ($\tau = 0.5$), the average suspect set contains approximately 1.94 candidates out of five (Table~\ref{tab:combined_results_table}). This indicates that participants tend to distribute probability mass across multiple plausible authors rather than concentrating belief on a single dominant candidate. Participants  also exhibit relatively medium confidence (mean = 0.49) despite limited accuracy, alongside weaker expected calibration error (ECE = 0.37), indicating a tendency toward uncertainty.

Overall, these results suggest that, when restricted to abstract-level information, human judgments exhibit limited selectivity and uncertainty, reflecting the difficulty of isolating authorship signals within semantically similar candidate sets, where distractors are drawn from domain-relevant authors. The relatively low MRR, together with its confidence interval, further suggests that this behavior is consistent across annotators and not driven by a small number of outliers.
 
\subsection{LLMs achieve efficient anonymity collapse}
We next evaluate whether large language models can improve upon human authorship inference under identical conditions. As in the human study, models are provided only with titles and abstracts and are required to rank and assign confidence scores over a fixed set of five semantically constrained candidate authors. 

Using the Qwen2.5-72B model, anonymity deteriorated. We observe a stronger concentration of probability mass on plausible candidates compared to human judgments with the true author frequently appearing within suspect set. The model achieves a top-1 accuracy of 42.34\%, significantly outperforming human performance of 11\% (Table~\ref{tab:topk_accuracy_table}).

The model outperforms human evaluators in ranking quality by achieving an MRR of 0.623 (95\% CI: [0.617, 0.629]) compared to 0.36 for humans. This improvement is statistically significant (Wilcoxon \(p < 0.001\)) with a large effect size (Cliff’s \(\delta = 0.52\)). The distributions of ranking scores differ significantly (KS test, \(p < 0.0001\)), indicating an upward shift in inference behavior. 

More importantly, at a moderate confidence threshold 50\% ($\tau = 0.5$), the model kept the average suspect set to 1.95 candidates while retaining the true author in 61.6\% of cases nearly 3 times higher than humans, demonstrating an ability to narrow the plausible author pool with better expected calibration error (ECE: 0.30) (Table~\ref{tab:combined_results_table}). This indicates that model probabilistic estimates are more aligned with empirical correctness, suggesting that model predictions are more reliable than human uncertainty.

Robustness checks confirm that these results are not artifacts of prompting or decoding. Removing confidence scoring and requiring the model to produce only a ranked list of candidates yields stable performance with a top-1 accuracy of approximately 40\% (Table~\ref{tab:topk_accuracy_table}), indicating that attribution performance does not depend on probability calibration or scoring. Shuffling candidate order and increasing decoding temperature from 0.0 to 0.3 produce no significant changes (Table~\ref{tab:topk_accuracy_table}), suggesting that it does not rely on positional biases or deterministic decoding.

Overall, these findings reveal a qualitative difference in inference strategy: humans distribute confidence across multiple plausible candidates, whereas language models concentrate probability mass more effectively, resulting in improved ranking and higher recall
\emph{defined as
the fraction of trials where the true author is included in the suspect set} at comparable levels of uncertainty. Operationally, these results show that even under idealized compliance with double-blind review guidelines, scientific text contains sufficient latent information to narrow authorship to a small, high-confidence subset differentiating between authors in the same field.

\subsection{Anonymity erosion persists across models but degrades under topical confounding}
Thirdly, we examine whether the observed anonymity collapse is model-specific and how it depends on the construction of distractor candidates. To this end, we evaluate independently trained large language models, including Qwen2.5-72B and LLaMA3-70B, under identical conditions: inputs restricted to titles and abstracts, and candidate pools of five authors, where distractors are domain experts working on closely related topics. We consider multiple strategies for constructing domain-expert distractors. In one setting, candidates are retrieved using Semantic Scholar recommendations \cite{ref3}, while in another, they are obtained via OpenAlex \cite{ref32}. Although both aim to identify topically similar authors, these sources differ in retrieval characteristics and candidate quality \cite{ref33}.

\begin{figure}[H]
\centering
\begin{minipage}{0.335\linewidth}
  \centering
  \includegraphics[width=\linewidth]{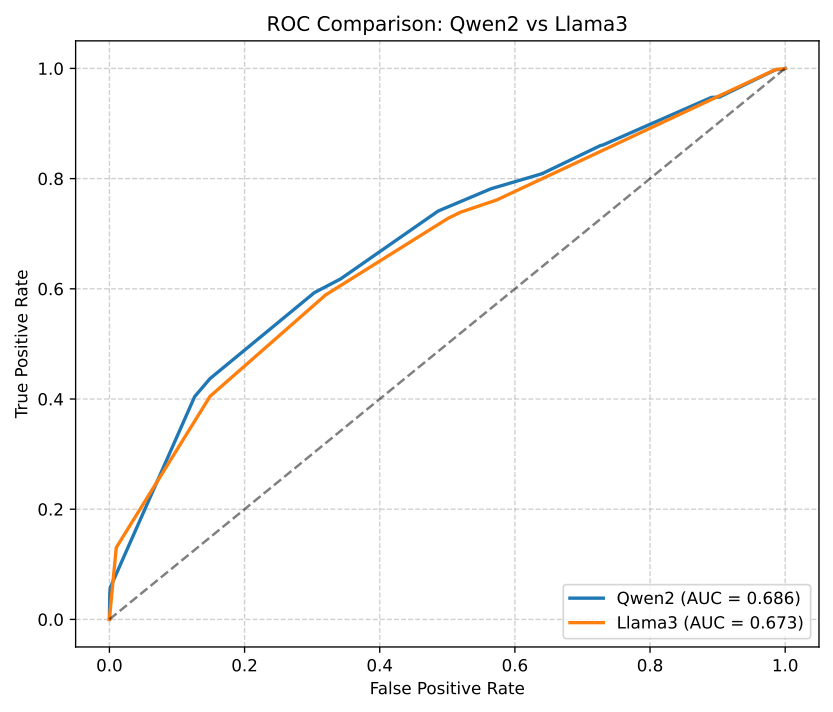}
  \caption{AUC-ROC for Qwen2 vs LLaMA3 on domain expert-level candidates.}
  \label{fig:model_auc}
\end{minipage}\hfill
\begin{minipage}{0.335\linewidth}
  \centering
  \includegraphics[width=\linewidth]{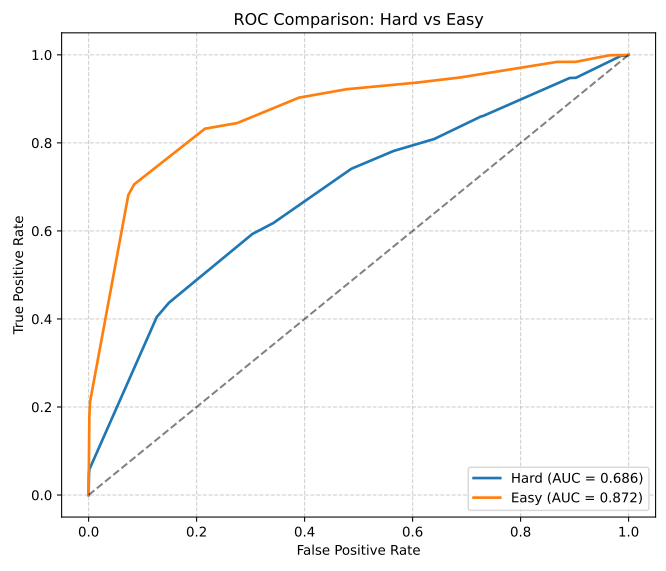}
  \caption{AUC-ROC comparison: random (Easy) vs domain-expert (Hard) distractors.}
  \label{fig:hard_easy_auc}
\end{minipage}\hfill
\begin{minipage}{0.29\linewidth}
  \centering
  \includegraphics[width=\linewidth]{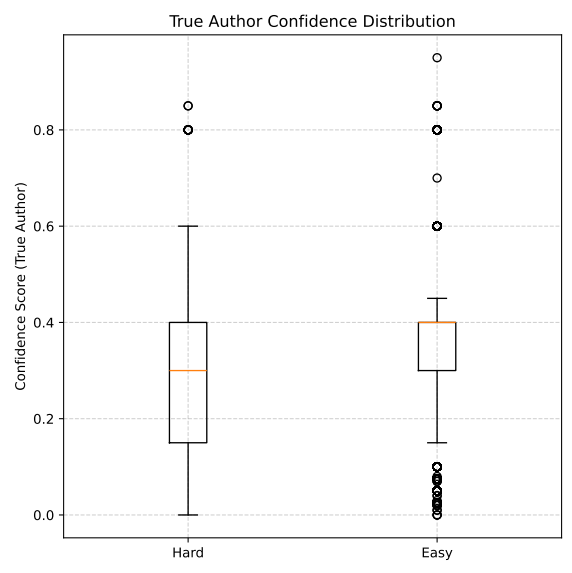}
  \caption{Distribution of suspect set sizes on random (Easy) vs domain-expert (Hard) distractors.}
  \label{fig:hard_easy_box}
\end{minipage}
\end{figure}

Across models, we observe consistent anonymity erosion as measured by the trade-off between recall and average suspect set size (Table~\ref{tab:combined_results_table}), with Qwen2.5-72B and LLaMA3-70B achieving close 
top-1 accuracy of approximately 42\% and 40\% respectively (Table.~\ref{tab:topk_accuracy_table})) and comparable discriminative performance AUC--ROC of approximately $0.67$ and $0.68$ respectively (Fig.~\ref{fig:model_auc})). Confidence intervals computed via bootstrap resampling (p > 0.1) indicate that these differences are not statistically significant, suggesting that anonymity erosion is not a model-specific artifact.

However, performance varies noticeably across distractor construction strategies. In the domain experts settings, the model’s predictions 
are less decisive, reflected in larger suspect sets, degraded performance and increased variability in confidence, suggesting difficulty distinguishing among candidates (Fig.~\ref{fig:hard_easy_auc}; Fig.~\ref{fig:hard_easy_box}).

In particular, distractors derived from OpenAlex yield more challenging candidate sets, reflected in only 51\% of recall with a larger suspect sets (Avg Set $\approx$ 2.78) at fixed confidence threshold $\tau = 0.5$ (Table~\ref{tab:combined_results_table}). This suggests that OpenAlex more reliably retrieves semantic related authors with closely overlapping research profiles, resulting in increased difficulty of distinguishing among authors. By contrast, distractors obtained via Semantic Scholar tend to produce medium candidate sets, reflected in 61.41\% recall and a bit smaller suspect sets (Avg Set $\approx$ 1.83) at the same confidence thresholds $\tau = 0.5$ (Table~\ref{tab:combined_results_table}). This is likely due to noisier or less strictly aligned retrieval, which can inadvertently introduce candidates that are less competitive or partially mismatched \cite{ref33}.

The overlap is confirmed via paired bootstrap tests (p < 0.01), indicating that OpenAlex produces significantly more confusable and competitive candidate sets. This difference is further reflected in calibration behavior. Under the OpenAlex setting, models exhibit higher Expected Calibration Error (ECE $\approx$ 0.082) compared to Semantic Scholar (ECE $\approx$ 0.057), indicating reduced alignment between predicted confidence and empirical correctness. Confidence distributions are also less sharp, with greater variance across samples, suggesting increased uncertainty when distinguishing among highly similar candidates.

The effect becomes most pronounced when candidate pools are randomly sampled from unrelated domains, where models can more easily eliminate implausible authors and concentrate probability mass on the true author. In this setting, anonymity collapsed rapidly such as at a similar moderate confidence threshold $\tau = 0.5$, the model narrowed the suspect set more aggressively from 2.78 to 1.82, including the true author in much higher 84\% of cases (Recall = 0.841) (Table~\ref{tab:combined_results_table}). This pattern is statically significat (p < 0.001), indicating that the model relies largely on topical alignment between the manuscript’s scientific content and the known research profiles of candidate authors.

Taken together, these findings indicate that anonymity erosion is not a model-specific artifact, but depends critically on the quality and confusability of the candidate set. While high-quality, topically aligned distractors attenuate performance, they do not eliminate the effect, suggesting that authorship signals persist even under carefully constructed, semantically similar alternatives.

\subsection{The strength of anonymity erosion varies systematically across disciplines}
Next, we checked whether the extent of anonymity collapse depends on disciplinary context. Restricting evaluation to papers from distinct fields of study such as medicine and computer science, we observed differences in the level to which effective anonymity was reduced.

For medical papers, effective anonymity eroded more rapidly than for computer science papers, with
higher recall and more compact suspect sets at matched confidence thresholds. At ($\tau = 0.5$), the true author was retained within the suspect set in approximately
66\% of medical cases, compared with approximately 58\% for computer science, while maintaining
smaller average suspect set sizes of approximately 2 candindates (Fig.~\ref{fig:disciplineandseniority_comparison};). This difference persisted across confidence levels and was reflected in how the model distributed a higher proportion of probability mass on the correct author compared with distractors for medical abstracts.


These results indicate that latent authorship signals are not uniformly distributed across scientific domains. Field-specific research conventions and shared problem framings may amplify conceptual regularities that undermine anonymity. 

\subsection{Anonymity erosion is not driven by author seniority or prominence}

A natural concern is that apparent anonymity collapse may be driven primarily by author prominence rather than natural properties of scientific text. To test this possibility, we stratified evaluation instances by career stage, designating the true author as (i) the most senior author, operationalized as the individual with the highest publication count; and (ii) the most junior author, defined as the individual with the lowest publication count.

Anonymity was reduced at comparable rates in both strata. Differences in recall and suspect set size across confidence thresholds were minor, indicating that anonymity erosion is not confined to highly prolific or well-known researchers. When the confidence threshold was set to ($\tau = 0.5$), the true author remained within the model’s suspect set in roughly half of the evaluation instances for both senior and junior researchers ($\approx 54\%$ for senior versus $\approx 52\%$ for junior authors). At the same time, the model typically reduced the candidate pool from five possible authors to about two on average in both settings (Avg Set $\approx 2$) (Fig.~\ref{fig:disciplineandseniority_comparison}). 
 
Prominence and discipline-specific notion of authorship importance are therefore not a necessary conditions for deanonymization: minimal scientific text alone suffices to narrow authorship across career stages.

\subsection{Increased inference complexity does not improve attribution}

Finally, we examined whether more complex inference strategies improve the model’s ability to identify the true author. Specifically, we evaluated (i) model aggregation, where predictions from multiple models are combined, and (ii) multi-round deliberation, where a model such as Qwen2.5-72B iteratively refines another model's (Llama3-70B) predictions.

Aggregating predictions across models yields only marginal improvements over the strongest single model (Qwen2.5-72B). As shown in Table~\ref{tab:combined_results_table}, both recall and average suspect set size remain nearly unchanged under aggregation. At a moderate confidence threshold ($\tau = 0.5$), both aggregated model and the strongest single model achieve a recall of $\approx 0.61$ with an average suspect set size of 1.88 and 1.95 respectively. Similar patterns hold across thresholds, indicating that different models tend to assign high probability to the same candidates and make similar errors.

In contrast, multi-round deliberation reduces attribution performance. From the second round of reasoning, the model assigns more similar probabilities across multiple candidates instead of concentrating probability mass on one or two authors. This leads to larger suspect sets and lower recall at fixed confidence threshold (Table~\ref{tab:combined_results_table}). For example, at confidence threshold $\tau = 0.5$, recall decreases from $\approx 0.61$ (Qwen2.5-72B) to $\approx 0.54$ under multi-round deliberation, while the average suspect set size increases from 1.95 to 2.22. The degradation becomes more pronounced at higher confidence thresholds: at $\tau = 0.8$, the average suspect set expands from 2.97 (Qwen2.5-72B) to 3.84 under debate, indicating substantially weaker concentration of probability mass. These results show that when the input is restricted to short scientific descriptions such as title and abstract and candidate authors have similar research profiles, the available authorship signal is limited. The second reasoning round does not introduce new information relevant to authorship.

\section{Discussion}

This study investigates whether large language models can reduce anonymity in double-blind review settings \cite{ref5, ref25}. Unlike previous work \cite{ref4, ref6, ref19, ref20, ref22, ref23} that exploit stylistic patterns, lexical regularities, and especially bibliographic and citation structures to recover authorship, our work does not extract and use such features. Given only titles and abstracts, across all experiments, we observe that models consistently reduce uncertainty over authorship beyond human intuition (Table~\ref{tab:combined_results_table}).

\begin{figure}[H]
  \centering
  \includegraphics[width=\linewidth]{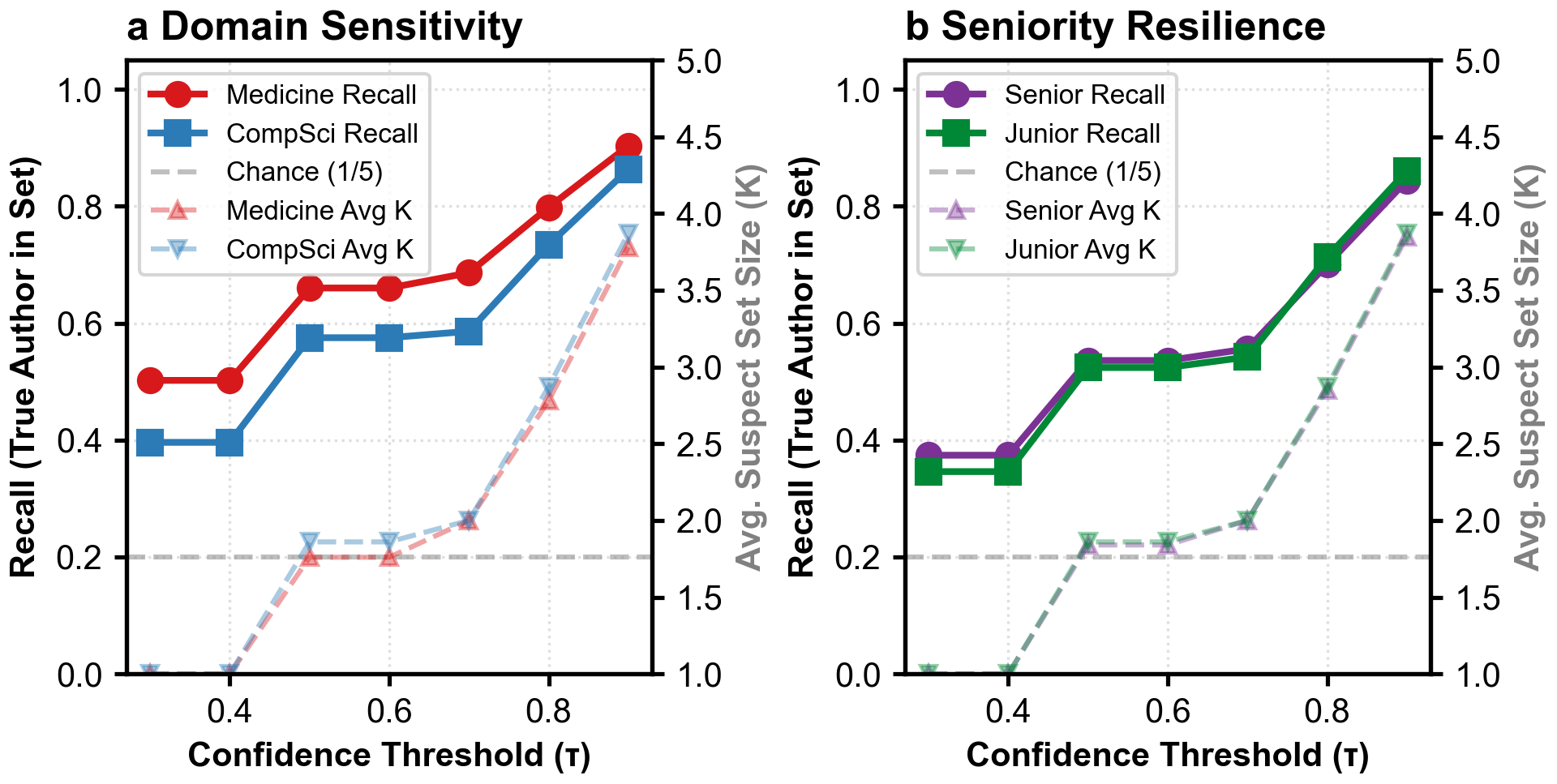}
  \caption{Systematic collapse of double-blind anonymity across scientific domains and career stages. Avg K denotes the average number of candidates in the suspect set.} 
  \label{fig:disciplineandseniority_comparison}
\end{figure}

The results do not suggest that double-blind review fails in an absolute sense. Instead, they reveal a graded vulnerability: even when explicit identifiers are excluded and candidate authors are closely matched by topic, title and abstract, information remains that allows automated systems to narrow authorship hypotheses. This demonstrates that anonymity is weakened not through exact identification, but through systematic narrowing of plausible authors. The key observations of this study are as follows:

\begin{itemize}
\item \textbf{Latent authorship information in scientific framing}: We define authorship signal as \emph{information contained in the scientific content, such as problem formulation, methodological choices, and domain focus, that allows a model to distinguish between candidate authors without relying on explicit identifiers such as writing style or citations}. This effect is directly supported by the comparison between candidate pools (Table~\ref{tab:combined_results_table}).
At a fixed confidence threshold $\tau = 0.5$, the model retains the true author in 84.1\% of cases when candidate authors are randomly sampled from unrelated domains. However, when candidates are restricted to domain experts working on similar problems, performance drops substantially to 61.6\% using distractors from Semantic Scholar and 51.5\% using distractors from OpenAlex. Despite these challenging conditions, model performance remains well above the 22.0\% cases in which human retained the true author. This contrast is particularly revealing: human researchers, despite their domain expertise, achieved a lower ranking accuracy, suggesting that authorship signals in minimal scientific text are invisible to human judgment but accessible to language models. Notably, the model’s predictions appeared to be driven primarily by inferred research focus and problem framing rather than superficial stylistic cues as the anonymity erosion is sensitive to similarity level between candidate authors’ research domains.
For example, two researchers working in {\it machine learning} may both study {\it classification}, but differ in emphasis such as robustness versus efficiency. These differences are reflected in how problems are framed in the abstract, allowing the model to distinguish between them. This challenges the core assumption underlying double-blind review: removing names, citations, and stylistic cues is sufficient to preserve anonymity \cite{ref5, ref25}. Even when these signals are excluded, the scientific content itself encodes identifiable authoship signals. As a result, anonymization procedures that focus only on surface-level features may be insufficient, since identity can be inferred from how research is framed rather than how it is written.

\item \textbf{Disciplinary variation}: We observe that the degree to which anonymity is reduced varies across disciplines. This pattern is evident in (Fig.~\ref{fig:disciplineandseniority_comparison} A;), where we compare the anonymity collapse across disciplines such medicine and computer science. The true author is included in the suspect set more frequently for medical papers than for computer science papers. A notable 8\% disparity emerged between the two disciplines, with medical abstracts exhibiting a significantly stronger authorship signal than those in computer science.
Another possibility is that medicine has much broader (semantically distant) sub-topics than in computer science making it easier to separate the probable 
authors given the title and abstract of a paper.
This suggests that the effectiveness of double-blind review is not uniform across fields raising concerns about unequal protection against bias across disciplines, meaning that some research communities may be systematically more exposed to identity leakage than others. As a consequence, the practical guarantees of double-blind review may vary systematically across disciplines rather than applying uniformly across science.

\item \textbf{Independence from visibility}: Importantly, anonymity reduction is not driven by author prominence. This is evident in the comparison between author's seniority. (Fig.~\ref{fig:disciplineandseniority_comparison} B;). When candidate pools are stratified by career stage, performance remains similar with the true author retained within the suspect set in approximately identical number of cases for both senior and junior authors settings at same moderate confidence threshold. This indicates that the model does not rely on external reputation, but instead extracts patterns directly from the content of the manuscript. This indicates that anonymity risks are not limited
only to well-known researchers. Even early-career or less-visible authors exhibit consistent patterns in how they frame problems, which can be exploited by models.

\item \textbf{Constraints on more elaborate inference}: 
 We compare the results obtained by single models to that obtained by
 models aggregation and multi-round predictions (Table~\ref{tab:combined_results_table}). Combining outputs from two models such as Qwen2.5-72B and Llama 3-70B, does not improve the results obtained by using only
 Qwen2.5-72B with the true author retained within the suspect set in approximately identical number of cases in both scenarios, implying that attribution errors are not distinctive to a particular model. This suggests that the the risk is not easily mitigated by restricting access to specific models. If the authorship signal is inherent in the content, then any sufficiently capable system may exploit it, making the challenge difficult to eliminate.
 Conversely, a boundary emerged during multi-round deliberation: rather than sharpening the model's focus, additional reasoning led to a 7.5\% decline in number of cases in which the true author is retained within the suspect set and a marginal expansion of the suspect set (Table~\ref{tab:combined_results_table}).
 Reasoning seems to introduce a bit of hesitancy on the part of the LLMs. 
This behavior suggests that second reasoning round does not introduce new information relevant to authorship. Instead, it increases uncertainty by redistributing probability mass across candidates rather than sharpening distinctions between them. Together, these results delineate a clear boundary condition: with today's technology, anonymity erosion performance cannot be reliably mitigated through increased inference complexity, either through model aggregation or multi-step deliberation.

\end{itemize}

Overall, these findings indicate that double-blind anonymization does not fully remove authorship relevant information. Instead, it transforms authorship inference into a probabilistic process, where models cannot always identify the correct author but can consistently reduce the set of plausible candidates. From the perspective of peer review, this reduction in uncertainty may be sufficient to influence expectations about authorship and potentially reviewer behaviour. Our results strengthen the idea
that double-blind review should be understood as a probabilistic safeguard rather than a guarantee of anonymity. While anonymization removes explicit identifiers, it does not eliminate the authorship signal that can reduce uncertainty when analyzed by modern language models. This raises fundamental questions about whether current anonymization practices \cite{ref5, ref25} remain adequate in an AI-augmented research ecosystem. By highlighting these emerging vulnerabilities, we suggest that the research community revisit and update reviewer guidelines and anonymization policies to address these risks.

\subsection{Limitations and avenues for future work}
Finally, we must acknowledge several limitations of our work. The analysis considers only two large open-weight models, and performance characteristics may differ for other architectures or proprietary systems such as GPT-4o or Gemini. Candidate sets are fixed at five, leaving open the question of how authorship signals behave as the pool of plausible candidates grows. The human evaluation was conducted with graduate-level researchers, and it is plausible that more experienced senior scholars with deeper familiarity with specific research communities may achieve higher attribution accuracy, potentially narrowing the performance gap with language models. The remaining anonymity despite using LLMs to deanonymize papers may be enough of a deterrent so as not to change the final outcomes of peer-reviewed articles; more research is needed in this area. Future research should examine how evaluator seniority and domain familiarity moderate human attribution performance, explore alternative formulations of candidate selection and evaluate whether similar effects persist when additional manuscript sections such as introduction or conclusion are included.

\section{Methods}

\subsection{Study Design}

We designed a controlled authorship attribution experiment to test whether large language models can reduce author anonymity using only minimal scientific text. The central objective was to isolate the authorship signals while excluding all explicit identifiers, stylistic cues, and bibliographic metadata. To this end, models were provided only with paper titles and abstracts and asked to reason over a small set of five plausible candidate authors. For example, a model might receive the title “Neural Scaling Laws for Multimodal Models” and an abstract describing training dynamics, together with five candidate authors such as “A. Smith, B. Lee, C. Zhang, D. Patel, E. Müller,” and must infer which is most likely to have written the paper.

Preliminary experiments using unconstrained generation, where models freely named likely authors based on a title and abstract, produced sparse and inconsistent outputs that did not support reliable statistical analysis. For instance, models often generated authors not present in the dataset or produced incomplete answers such as a single name without justification, making systematic evaluation infeasible.

We therefore adopted a confidence-aware ranking framework that measures how models distribute belief across a constrained set of five candidate authors, allowing us to quantify how rapidly anonymity collapses as confidence increases. Rather than framing authorship inference as a prediction task that requires the model to assign exactly one author as the correct output for each input instance, we evaluate whether the true author appears within a dynamically sized suspect set constructed by accumulating model-assigned belief until a predefined confidence threshold is reached. For example, if a model assigns to each of the five candidates confidence scores of [0.40, 0.30, 0.15, 0.10, 0.05], then at threshold $\tau$ = 0.5 the suspect set would include the top two candidates (0.40 + 0.30 $\ge$ 0.5).

This framing reflects realistic adversarial use: the risk to double-blind review lies not in perfect identification, but in the systematic reduction of plausible authorship to a small, high-confidence subset.

\subsection{Dataset construction and temporal isolation}

We constructed a corpus of scientific papers published after the models’ documented training cutoff
date to test whether the model can handle new, unseen papers, rather than just remembering
the authors of 
papers it has already seen during training. Roughly 25,000 papers were collected from the Semantic Scholar API, spanning broad fields of study and dated between January 2024 and October 2025.

To ensure evaluation samples were not implicitly present in pretraining data, all candidate papers were cross-referenced against the arXiv preprints corpus through December 31, 2023. Titles or abstracts that matched earlier preprints under fuzzy string matching were removed. For example, a 2024 journal paper that had an earlier 2023 arXiv version with highly similar text would be removed. 
We note that our contamination filtering primarily targets preprints indexed in arXiv and may miss earlier versions hosted on alternative platforms (e.g., bioRxiv, SSRN, or personal webpages). To assess the extent of such residual contamination, we conducted a manual audit of a random sample of 200 papers from the filtered dataset. For each paper, we performed targeted Web searches using title and author queries to identify potential earlier versions. We found that 0.5\% of sampled papers had evidence of pre-2023 publicly available versions not captured by our automated filtering. The observed differences were minimal, suggesting that our findings are robust to such contamination.
This filtering step ensured that retained samples were both temporally and textually novel relative to the models’ training data, providing a stringent test of generalization rather than memorization.

For each paper, we retained only five fields: a unique paper identifier, title, abstract, fields of study, and the full author list. All other metadata were discarded to eliminate additional identifying signals.
 
\subsubsection{Discipline Specific datasets}

To reduce cross-field contamination, we selected papers assigned to only a single field of study. For instance, a paper labeled only as “Computer Science” was retained, whereas one labeled “Computer Science” and “Physics” was excluded. From these, we curated subsets in Medicine and Computer Science for domain-specific analyses.

\subsection{Candidate selection and topical confounding}

\subsubsection{Ground thruth Candidate selection}

Three candidate sets were constructed to reflect authorship roles. We consider multiple definitions of the ground-truth author to account for variability in authorship conventions across disciplines. Specifically, we evaluate three target authors for each paper: (i) the last author, used as our primary baseline; For example, in a paper with authors [A, B, C, D], author D would be treated as the true author across all experiments. (ii) the most senior author, operationalized as the individual with the highest publication count; and (iii) the most junior author, defined as the individual with the lowest publication count. For instance, in the senior-author setting, the true author is selected from the actual paper’s author list as the individual with the highest publication count. In contrast, the junior-author setting selects the true author as the least-published co-author, representing an early-career researcher. This design ensures that our findings are robust and not tied to a single, potentially discipline-specific notion of authorship importance.

\subsubsection{Topical confounding}
We define higher topical confounding as a setting in which candidate authors exhibit substantial overlap in research topics, thereby reducing the discriminative power of topical features. To probe model behavior under varying degrees of ambiguity, we constructed three distratctors collection regimes that differ in the degree of topical confounding. In all cases, candidate sets consisted of four distractors not listed on the actual paper's authors list.

In the first regime, distractors were sampled from a broad pool of researchers drawn across diverse academic fields obtained via the Semantic Scholar API \cite{ref31}.

In the second, more stringent regime, distractors were restricted to topically similar domain experts authors recommended by the Semantic Recommendation API \cite{ref3}. For each paper, we used recommendation based retrieval to identify related publications and collected authors from these thematically aligned papers to form a pool of plausible in-field candidates. Prior work has shown that such recommendation systems capture semantic similarity but may introduce noise or leakage artifacts due to retrieval biases or imperfect citation-based signals \cite{ref33}. 

To mitigate these limitations, we additionally construct a second expert distractor set using OpenAlex \cite{ref32}, which provides conceptually grounded and structurally consistent author–publication linkages. This enables a more controlled comparison of authorship inference under two complementary notions of topical similarity.

In the broad setting, a machine learning paper might be paired with candidates from biology or economics. In the expert setting, all candidates would be machine learning researchers working on similar topics such as transformers or representation learning. All Candidate lists were randomly shuffled before evaluation to prevent position bias.

\subsection{Models and execution environment}

Experiments were conducted using two open-weight large language models with documented 2023 training cutoffs: Qwen2.5-72B \cite{ref1} and Llama 3-70B \cite{ref2}. These models were selected to enable fully controlled local deployment, ensuring that all predictions arise solely from the models’ internal representations without access to external data sources. This design eliminates confounding factors such as retrieval augmentation, web search, or API-based knowledge injection.

We deliberately do not evaluate larger proprietary systems (e.g., GPT-4, Claude or Gemini models), as such systems may implicitly rely on retrieval-augmented generation (RAG) or access to continuously updated external knowledge, making it difficult to disentangle memorization from inference. In contrast, our setup isolates the intrinsic capability of language models to infer authorship from minimal textual signals under strictly offline conditions.

Additionally, our experiments were conducted under practical infrastructure constraints that limit the deployment of larger-scale models beyond the 70B parameter range. Despite this, the selected models represent the upper bound of what can be reliably executed in a fully controlled, offline environment. As such, our results should be viewed as a conservative estimate: more capable systems may further amplify deanonymization performance.

\subsection{Prompting and inference}

For each evaluation instance, models were provided with the paper title and abstract, together with a shuffled list of candidate author names. Models were instructed to rank all candidates from most to least likely and to assign a confidence score to each candidate.  To ensure reproducibility, the exact prompt provided to the models is shown below:

\begin{verbatim}
You are an expert research analyst simulating a peer reviewer.
Given:
- Paper title
- Paper abstract
- A list of 5 candidate authors
Task:
1. Rank all candidates from most to least likely author
2. Assign a confidence score to each
Output ONLY valid JSON:
{
  "ranked_authors": [
    {"name": "string", "confidence_score": float}
  ],
}
\end{verbatim}



Confidence scores were interpreted as relative belief assignments and were accumulated over the ranked list to construct suspect sets at varying confidence thresholds.

\subsection{Evaluation metrics and statistical analysis}

We evaluated model performance using complementary metrics designed to capture authorship attribution accuracy, ranking quality, discriminatory power and confidence concentration.

Top-k accuracy measures whether the true author appears within the top k ranked candidates. This evaluates coarse-grained retrieval performance independent of probability calibration. We also report Mean Reciprocal Rank (MRR) to capture ranking quality across the full candidate list:
\[
\mathrm{MRR} = \frac{1}{N} \sum_{i=1}^{N} \frac{1}{\mathrm{rank}_i}
\]
where rank i denotes the position of the true author in the ranked candidate list. Higher values indicate stronger prioritization of the correct author. To quantify uncertainty in ranking estimates, we compute 95\% confidence intervals (CI) for MRR using bootstrap resampling over test instances.

Suspect set recall measures whether the true author appears within the suspect set at a given confidence threshold $\tau$. Suspect set size quantifies the average number of candidates required to reach that threshold, providing a measure of how efficiently models reduce anonymity.
To assess ranking quality independently of threshold choice, we computed micro-averaged area under the receiver operating characteristic curve (AUC–ROC) in a one-vs-rest formulation. Values substantially above indicate consistent prioritization of the true author over distractors.

To evaluate how well the models' self-reported confidence scores align with empirical accuracy, we calculated Expected Calibration Error (ECE). This metric quantifies the difference between predicted confidence and observed accuracy, where lower values indicate better-calibrated models. 

Statistical Significance Testing includes the Wilcoxon signed-rank test used to compare the ranking quality (MRR) between human and model performance.  Cliff’s $\delta$ a non-parametric effect size measure used to quantify the magnitude of difference between human and model inference.  Kolmogorov-Smirnov (KS) test is employed to determine if the distributions of ranking scores between different groups (e.g., humans vs. models) differ significantly.  Paired bootstrap tests used to compare performance across different distractor construction strategies (e.g., Semantic Scholar vs. OpenAlex). 

We additionally examined the distribution of suspect set sizes and recall across batches of test papers to assess variability under different levels of topical confounding. 
All inference was performed using a parallelized evaluation pipeline to ensure consistency across datasets, models, and experimental conditions.

\begin{table}[ht]
\centering
\caption{Top-$k$ accuracy (\%) across human, models 
and candidate selection strategies.}
\label{tab:topk_accuracy_table}
\tiny
\resizebox{\textwidth}{!}{%
\begin{tabular}{lcccc}
\hline
\textbf{Model / Setting} & 
\textbf{Top-1 (\%)} & 
\textbf{Top-2 (\%)} & 
\textbf{Top-3 (\%)} & 
\textbf{Top-4 (\%)} \\
\hline
Human Researcher       & 11.00 & 23.00 & 41.00 & 61.00 \\
LLaMA3 OpenAlex Expert & 36.97 & 54.16 & 69.06 & 83.39 \\
LLaMA3 Semantic Expert & 40.41 & 59.70 & 74.81 & 87.79 \\
Qwen2 Semantic Expert  & 42.34 & 61.72 & 76.44 & 88.24 \\
Qwen2 Random           & 68.53 & 84.47 & 91.70 & 95.81 \\
Models Aggregation     & 41.84 & 62.10 & 76.73 & 88.91 \\
Models Debate          & 35.32 & 46.41 & 61.53 & 81.30 \\
LLaMA3 No Scores       & 40.06 & 59.62 & 74.47 & 87.56 \\
LLaMA3 Shuffled Author & 39.50 & 58.99 & 74.32 & 87.00 \\
\hline
\end{tabular}%
}
\end{table}

\begin{table}[H]
\centering
\caption{Deanonymization performance under different 
experimental conditions. Recall and average suspect 
set size (Avg Set) are reported for multiple confidence 
thresholds ($\tau$).}
\label{tab:combined_results_table}
\resizebox{\textwidth}{!}{%
\begin{tabular}{ccccccccccccccc}
\hline
& \multicolumn{2}{c}{\textbf{Human}}
& \multicolumn{2}{c}{\textbf{LLaMA3 OpenAlex}}
& \multicolumn{2}{c}{\textbf{LLaMA3 Semantic}}
& \multicolumn{2}{c}{\textbf{Qwen2 Semantic}}
& \multicolumn{2}{c}{\textbf{Qwen2 Random}}
& \multicolumn{2}{c}{\textbf{Aggregation}}
& \multicolumn{2}{c}{\textbf{Debate}} \\
\textbf{$\tau$} 
& Recall & Avg 
& Recall & Avg 
& Recall & Avg 
& Recall & Avg 
& Recall & Avg 
& Recall & Avg 
& Recall & Avg \\
\hline
0.30 & 0.130 & 1.30 & 0.370 & 1.00 & 0.404 & 1.00 
     & 0.424 & 1.00 & 0.686 & 1.00 & 0.431 & 1.05 
     & 0.386 & 1.00 \\
0.40 & 0.180 & 1.68 & 0.370 & 1.00 & 0.404 & 1.00 
     & 0.445 & 1.09 & 0.692 & 1.03 & 0.545 & 1.51 
     & 0.392 & 1.43 \\
0.50 & 0.220 & 1.94 & 0.515 & 1.78 & 0.577 & 1.83 
     & 0.616 & 1.95 & 0.841 & 1.82 & 0.615 & 1.88 
     & 0.541 & 2.22 \\
0.60 & 0.310 & 2.40 & 0.515 & 2.78 & 0.577 & 1.83 
     & 0.616 & 1.95 & 0.841 & 1.82 & 0.655 & 2.15 
     & 0.561 & 2.82 \\
0.70 & 0.390 & 2.74 & 0.542 & 2.00 & 0.597 & 2.00 
     & 0.637 & 2.14 & 0.852 & 2.10 & 0.729 & 2.61 
     & 0.602 & 3.10 \\
0.80 & 0.440 & 3.24 & 0.678 & 2.80 & 0.739 & 2.84 
     & 0.766 & 2.97 & 0.917 & 2.84 & 0.804 & 3.20 
     & 0.711 & 3.84 \\
0.90 & 0.680 & 4.02 & 0.823 & 3.80 & 0.872 & 3.84 
     & 0.880 & 3.92 & 0.957 & 3.79 & 0.887 & 3.92 
     & 0.807 & 4.11 \\
\hline
\end{tabular}%
}
\end{table}

\bibliographystyle{plain}
\bibliography{references}

\subsection{Acknowledgements}
Acknowledge Carnegie Mellon University Africa for institutional support, Semantic Scholar and OpenAlex APIs for data access, and the twenty human evaluators who participated in the study.

\subsection{Author contributions}
Bulambo Mwendelwa Gloire (B.M.G) and Prasenjit Mitra (P.M.) conceived and designed the study. B.M.G. developed the experimental framework, constructed the evaluation corpus, implemented the attribution pipeline, and conducted all computational experiments. B.M.G. performed the statistical analyses with input from P.M. B.M.G. designed and administered the human evaluation study. B.M.G. and P.M. interpreted the results. B.M.G. wrote the manuscript with critical revision and intellectual input from P.M. P.M. supervised the project and acquired funding. All authors read and approved the final manuscript.

\subsection{Materials and Correspondence}
\subsubsection{Data availability}
To support reproducibility, we release the raw corpus of scholarly papers collected via the Semantic Scholar API, which serves as the sole source of data for all experiments. All derived datasets, including the \emph{Randomized}, \emph{Domain-Expert}, \emph{Seniority} and \emph{Disciplines} pools, are generated deterministically from this base corpus following the procedures described above. The scripts used for dataset construction, along with the exact LLM prompting templates, are provided in the supplementary materials to enable independent replication.

\subsubsection{Code availability}
All code used for data processing, evaluation, and analysis is publicly available at:
\url{https://github.com/anonymous-authorship-collapse/unmask-double-blind-anonymisation}.



\subsubsection{Ethics statement}
Human evaluation participants were graduate-level researchers who provided informed consent prior to participation. No personally identifiable information was collected from participants beyond institutional affiliation and research expertise. The study did not involve sensitive personal data, clinical procedures, or vulnerable populations.





\end{document}